\documentclass{article}
\pdfoutput=1
\usepackage{ijcai22}

\usepackage{times}
\usepackage{xr}
\usepackage{soul}
\usepackage{url}
\usepackage[subpreambles=true]{standalone}
\usepackage[hidelinks]{hyperref}
\usepackage[utf8]{inputenc}
\usepackage{caption}
\usepackage{float}
\usepackage{graphicx}
\usepackage{stfloats}
\usepackage{amsmath}
\usepackage{import}
\usepackage{amsthm}
\usepackage{multirow}
\usepackage{booktabs}
\usepackage{multirow}
\usepackage{algorithm}
\usepackage{algorithmic}
\title{OneDConv: Generalized Convolution For Transform-Invariant Representation}

\author{
Tong Zhang\and
Haohan Weng\and
Ke Yi\and
C. L. Philip Chen
\affiliations
South China University of Technology
\emails
 tony@scut.edu.cn,
\{cswhaohan, 201930344343\}@mail.scut.edu.cn,
philipchen@scut.edu.cn
}

\begin{document}

\maketitle

\begin{abstract}
Convolutional Neural Networks (CNNs) have exhibited their great power in a variety of vision tasks. However, the lack of transform-invariant property limits their further applications in complicated real-world scenarios. In this work, we proposed a novel generalized one dimension convolutional operator ({\it OneDConv}), which dynamically transforms the convolution kernels based on the input features in a computationally and parametrically efficient manner. The proposed operator can extract the transform-invariant features naturally. It improves the robustness and generalization of convolution without sacrificing the performance on common images. The proposed OneDConv operator can substitute the vanilla convolution, thus it can be incorporated into current popular convolutional architectures and trained end-to-end readily. On several popular benchmarks, OneDConv outperforms the original convolution operation and other proposed models both in canonical and distorted images.
\end{abstract}

\section{Introduction}

Convolutional Neural Networks (CNNs) can extract expressive learning representations from high-dimensional data, achieving amazing performance on current visual benchmarks \cite{liu2022convnet}. The weight sharing mechanism, in which the convolution filter parameters are shared across all spatial positions, helps to extract the common features regardless of how the input images are translated. Nevertheless, current convolutional models are still ineffective in tackling other transformations like rotation and reflection. With the lack of an internal mechanism to deal with affine transformations, CNNs are brittle to generalize to a variety of poses of real-world objects. An enormous number of replicated feature detectors or labelled training images are required exponentially as the dimensions of affine transformations increase \cite{hinton2011transforming,sabour2017dynamic}.

The most common method for alleviating this problem is Data Augmentation \cite{van2001art}, which automatically generates predefined types of transformed images for training. But in this case, more network parameters and training efforts are required to guarantee the recognition of redundant patterns, even though there is only a minor transformation. A more elegant solution is to endow the convolution with the capability of transformation invariance. Many learnable convolution filters \cite{cohen2016group,shen2016transform,zhou2017oriented,xu2020towards} have been proposed recently to enhance the robustness against transformations, improving the performance on distorted images. 

However, on the one hand, strong assumptions about transformation types should be predefined, but the types cannot be predicted ahead of time in practice. For instance, \cite{cohen2016group} is proposed for a group of symmetries like rotation and reflection, while \cite{zhou2017oriented} is designed especially for rotation. These filters \cite{shen2016transform,xu2020towards}, on the other hand, focus and investigate only the performance on distorted images. It is insufficient since the real-world dataset contains both canonical and deformed images, and the performance on all these images should be promised.

In this paper, we propose a generalized convolution operator called {\it OneDConv}, to improve the transform-invariant ability against a variety of transformations without sacrificing performance on canonical images. The $n \times n$ square kernel is separated into n $1 \times n$ one-dimensional filters, and the distance between these one-dimensional kernels is determined dynamically during inferring. OneDConv is simple to implement in the current deep learning framework and fully utilizes GPU acceleration for convolution computation. For canonical images, the dynamical filter approaches the square filter, which is identical to vanilla convolution. In the case of distorted images,  OneDConv transforms itself to match the distortion of the input images. From the perspective of the internal mechanism, the convolution filters are endowed with the transformation-invariant capability, using only a few additional parameters in a computationally efficient way.

The proposed OneDConv is evaluated on both original and distorted images on MNIST, CIFAR10, and ImageNette. Several state-of-the-art learning networks are compared, and the comprehensive results show that OneDConv is promised to be valuable yet challenging work. Our model can achieve comparable results on canonical images while improving the performance on distorted images by over $3\%$ on accuracy.

\section{Related Work}

\subsection{Handcrafted Invariant Descriptors}

Traditional image descriptors use a series of handcrafted filters based on prior knowledge to represent important image regions, such as Gabor features \cite{gabor1946theory}, SIFT (Scale Invariant Feature Transform) \cite{ng2003sift} and LBP (Local Binary Patterns) \cite{ahonen2006face}. These local features are based on handcrafted descriptors like texture and histograms that are usually resistant to image transformations like rotation and translation. The major limitation of these methods is that their features may not always be able to adapt to real-world tasks and it is difficult to utilize a large amount of existing data. 

\subsection{Data Augmentation}

Data augmentation is a useful technique for increasing the size of the training dataset by incorporating transformed versions of the original images \cite{van2001art}. It generates predetermined sorts of transformed images automatically and feeds them into the network for training. TI-pooling \cite{laptev2016ti} employs a parallel network design to extract image features from the different orientations, as well as a global pooling structure before the top classification layer to ensemble diverse orientations of the features. 

Nevertheless, the models are required to remember all the patterns they have seen without exploring the relations between them, so more parameters and training cost are needed to recognize all these patterns.

\subsection{Transform-Invariant Module}

Several studies on CNNs \cite{jaderberg2015spatial,li2020feature,xu2020towards} have attempted to solve the problem of learning transformation-invariant representations by proposing plug-and-play modules to boost the invariant ability. Spatial Transformer Network (STN) \cite{jaderberg2015spatial} introduces a localisation module that automatically warps distorted images into canonical ones, showing success in small-scale image classification tasks. Feature Lenses \cite{li2020feature} are introduced to counteract various image transformations towards invariant image representations. \cite{xu2020towards} propose a multi-scale max-out block as well as a regulator.

For these methods, they incorporate the specially designed structures into the original convolution network architecture, which does not address the problem of convolution's inherent lack of transformation invariance.

\subsection{Transform-Invariant Filter}

Many efforts appear to improve the transformation-invariant ability of convolutional filters. Active Rotational Filter (ARF) \cite{zhou2017oriented} is proposed to generate the features explicitly encoded with the location and orientation information. Each ARF spins and builds feature maps to capture the response of receptive fields in predefined directions. \cite{cohen2016group} proposed Group equivariant CNNs (GCNN) to utilize wider groups of symmetries, such as rotations and reflections. \cite{worrall2017harmonic} proposed Harmonic Networks by replacing regular CNN filters with circular harmonics and returning a maximal response and orientation for every receptive field patch. Deformable Convolution \cite{dai2017deformable} adds 2D offsets to the typical grid sampling sites of convolution, allowing the sample grid to be deformed. Deformable Convolution resembles our work in spirit, but it is designed specifically for semantic segmentation and object detection.Gabor Convolution \cite{luan2018gabor} incorporates Gabor filters into the convolution filter, reinforcing the robustness of learned features against the orientation and scale changing. 

These filters can achieve high accuracy on distorted images with the prior knowledge or assumptions brought in, but they are implemented in a complicated manner, and may affect the performance on origin images to some extent.

\section{Method}
\subsection{Overview}

\begin{figure}
    \centering
    \includegraphics[width=0.4\textwidth]{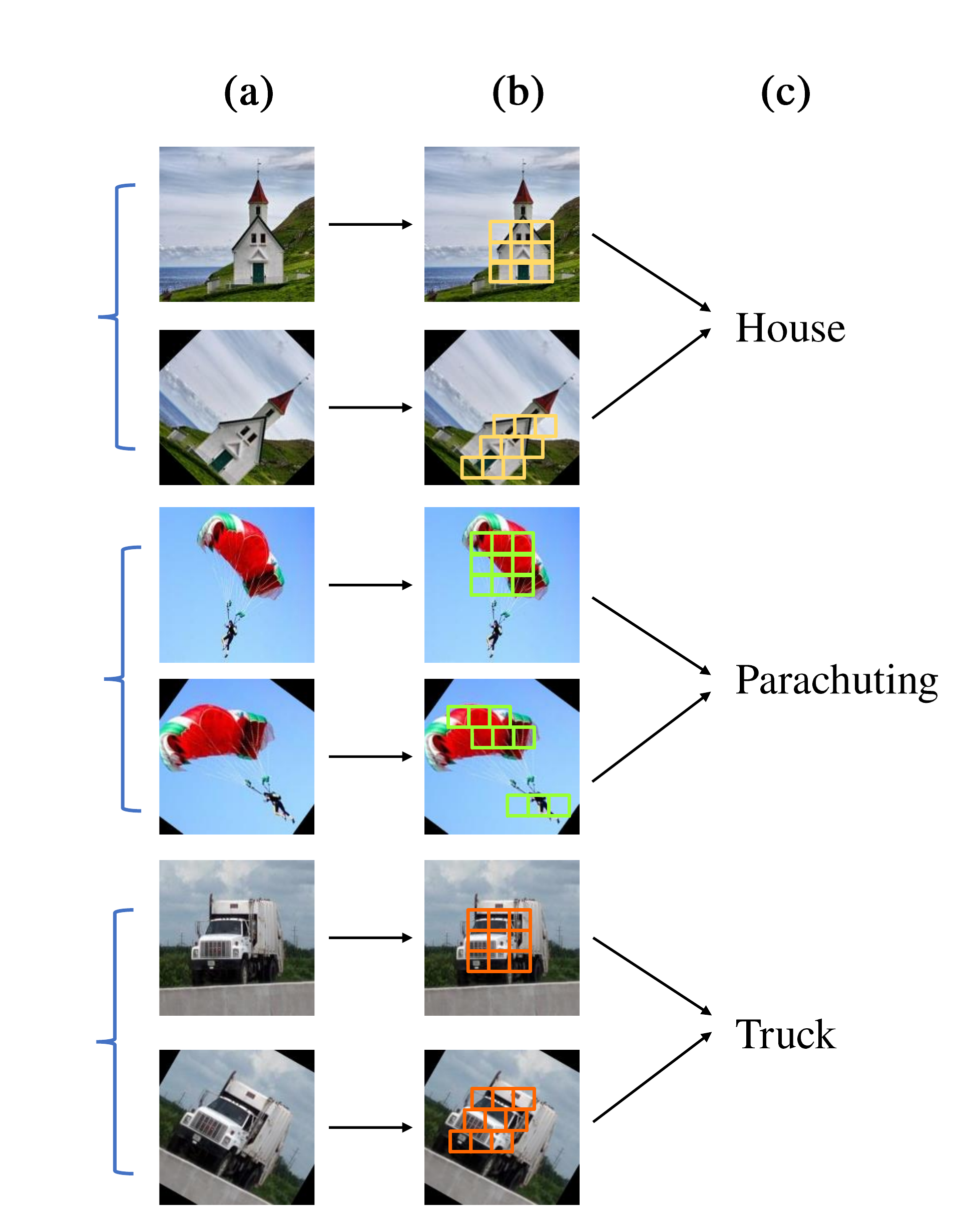}
    \caption{The procedure of using OneDConv in a Convolution Neural Network trained for robust classification  is described. (a) There are three image group. Each group has an origin image and its distorted version. (b) The dynamic filter adjusts itself to the distortion of the image. For canonical images, the dynamical filter approaches the square filter. (c) Predictions are generated by the network which is trained end-to-end without prior knowledge for the input distortion.}
    \label{fig:overview}
\end{figure}

To extract transform-invariant features, One Dimensional Convolutional operator (OneDConv) is proposed, which can be applied to different types of CNN architectures. The operator retains properties of convolution operators, and is generalized for further robustness by learning the shape of objects. Arbitrary networks containing convolution layers can directly use it to improve performance against all sorts of transformations.

A typical convolution filter can be separated into a series of one-dimension filters, which is the main idea of OneDConv. As shown in Figure \ref{fig:overview}, $3 \times 3$  filters are separated into three $1 \times 3$ filters, having a capability to cover the entire object in the feature map. Note that the shape approaches a square while inputting a canonical image. But for distorted images, OneDConv adjusts itself to match the distortion of the input images. 

Compared with dividing the $3 \times 3$ filter into nine $1 \times 1$ filters, it involves less overhead in terms of parameters and complexity while strengthening the receptive fields. Besides that, the proper compactness and flexibility enable the filters to learn transform invariant features. In the sequel we will explain how to design OneDConv and the shape of it.

\subsection{Shape Convolution}

This section will introduce a concept for the shape of a OneDConv filter. The distances between adjacent filters are used to parametrize the shape of a convolution filter. For example, assume that a OneDConv filter is sliding on a certain feature map. The filter is a composition of K $1 \times K$ one-dimension filters, and the height and width of the input feature map are $H \times W$. The starting location of each filter is denoted as $(L^i_x,L^i_y)$. The distance between each one-dimension filter can be expressed as
\begin{equation}
W-L^i_x+(L^{i+1}_y-L^i_y)\cdot w+L^{i+1}_x.    
\end{equation}
If all of the filters form a square like a 2D convolution filter, the distance between each filter will be a constant $W$. 

Rather than presetting the shape of OneDConv, a convolution layer named ShapeConv is proposed to learn the shape dynamically while training, which is formulated as
\begin{equation}
s=w_s \bigotimes I_{in}   \label{shape convolution}
\end{equation}%
where $w_s$ denotes the shape convolution filters, $I_{in}$ denotes the input feature map.
ShapeConv samples over the same input feature map while the output feature map has the same width and height as the input. Meanwhile, the number of output channels is set to the same as the filter quantity minus 1. As a result, each 2D position on the output feature map can represent a specific shape of an OneDConv filter by values on the channels.

Shape Convolution is viewed as the first part of OneDConv, its output will affect the shape of the latter convolution filter. Through backward propagation, the kernel of the Shape Convolution will be optimized without additional supervision. Details will be given in the next section.
\subsection{One-dimension Convolution}

\begin{figure}
\centering  
\includegraphics[width=0.4\textwidth]{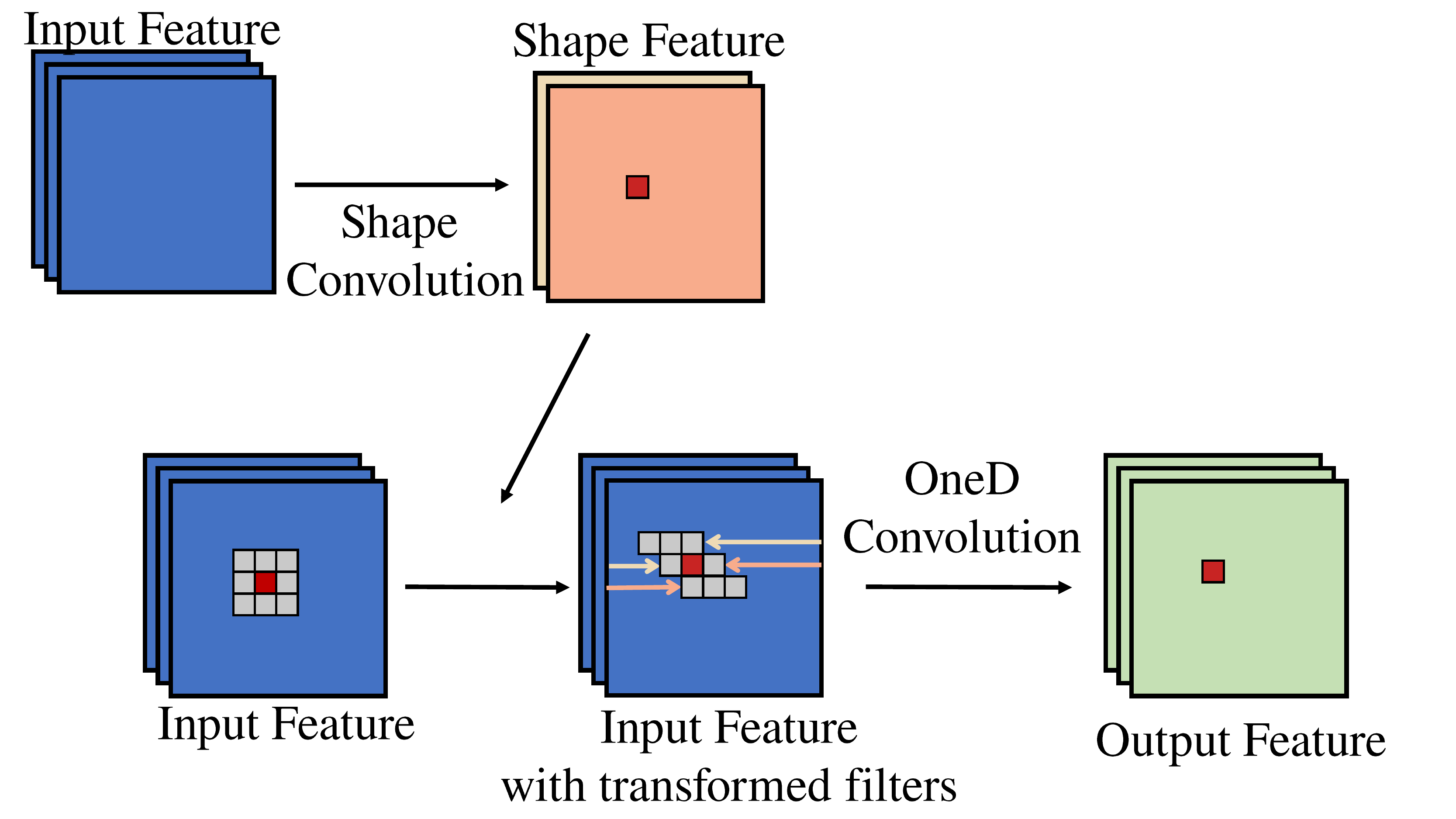}  
\caption{OneDConv learns the shape of filters dynamically. The shape feature is obtained by Shape convolution with $N - 1$ channels. The OneDConv filter transforms its shape each time it slides, depending on the shape feature when forwarding.}  
\label{fig:dynamic}
\end{figure}

\begin{table*}

\renewcommand\arraystretch{1.5}
\centering
\setlength{\tabcolsep}{1.7mm}{
\scalebox{0.9}{
\begin{tabular}{c|c|c|c}
\hline
layer name  & 18-layer & 34-layer & pyramid net 110\\
\hline
conv1  & \multicolumn{2}{c|}{$ OneDConv, 64,stride 2$} & $\left[ 3\times 3,16 \right]$ \\
\hline
conv2\_x  &$
\left[
\begin{matrix}
OneDConv, 64 \\
OneDConv, 64 \\
\end{matrix} \right] \times 2
$
&
$
\begin{matrix}
3 \times 3 max pool, stride 2\\
\left[
\begin{matrix}
1 \times 1, 64 \\
OneDConv, 64 \\
1 \times 1, 256\\
\end{matrix} \right] \times 2
\end{matrix}
$
&
$
\left[
\begin{matrix}
3\times 3,\lfloor 16 + \alpha(k-1)/N\rfloor\\
3\times 3,\lfloor 16 + \alpha(k-1)/N\rfloor\\
\end{matrix}
\right] \times 18
$
\\

\hline
conv3\_x &  $
\left[
\begin{matrix}
OneDConv, 128 \\
OneDConv, 128 \\
\end{matrix} \right] \times 2
$ & $
\left[
\begin{matrix}
1 \times 1, 128 \\
OneDConv, 128 \\
1 \times 1, 256
\end{matrix} \right] \times 4
$ & 
$
\left[
\begin{matrix}
3\times 3,\lfloor 16 + \alpha(k-1)/N\rfloor\\
3\times 3,\lfloor 16 + \alpha(k-1)/N\rfloor\\
\end{matrix}
\right] \times 18
$
\\
\hline
conv4\_x &  $
\left[
\begin{matrix}
OneDConv, 256 \\
OneDConv, 256 \\
\end{matrix} \right] \times 2
$ & $
\left[
\begin{matrix}
1 \times 1, 256 \\
OneDConv, 256 \\
1 \times 1, 1024
\end{matrix} \right] \times 6
$ & 
$
\left[
\begin{matrix}
3\times 3,\lfloor 16 + \alpha(k-1)/N\rfloor\\
3\times 3,\lfloor 16 + \alpha(k-1)/N\rfloor\\
\end{matrix}
\right] \times 18
$
\\

\hline
conv5\_x & $
\left[
\begin{matrix}
OneDConv, 512 \\
OneDConv, 512 \\
\end{matrix} \right] \times 2
$ & $
\left[
\begin{matrix}
1 \times 1, 512 \\
OneDConv, 512 \\
1 \times 1, 2048
\end{matrix} \right] \times 3
$ & None\\
\hline
&  \multicolumn{3}{c}{average pool. 10-d fc, softmax} \\
\hline

\end{tabular}}}

\caption{Network Architecture}
\label{tab:network}
\end{table*}

Forward propagation inner the 2D convolution consists of two steps: 1) sampling across a regular grid $R$ over the input feature map $x$; 2) summation of sampled values weighted by $w$. In OneDConv, the 2D convolution filter is substituted by a collection of one-dimension filters and feature maps are flattened. The grid $R$ defines the receptive field and dilation for each one-dimension filter. For example,$R=\{-1,0,1\}$ defines a $1 \times 3$ filter with dilation 1. 

For each location $l_0$ on the output feature map $y$ 
\begin{equation}
y(l_0)=\sum^{N}_{i=1}\sum_{l_n \in R}w_i(l_n) \cdot x(l_0+l_n+(i-\lceil N/2\rceil)\cdot W) \label{noraml convolution}
\end{equation}%

Here we define $N=|R|$, which denotes the number of one-dimension filters and also the size of a one-dimension filter, $l_n$ enumerates the relative locations in $R$, $W$ denotes the width of the input feature map. $w_i$  denotes the weight of ith one-dimension filter.

To augment the regular grid R, apply ShapeConv sampling over the input feature map $x$. The output feature map from Shape Convolution will also be flattened and represented as $d$, where $d_i$ is the ith channel of output.  
Eq. (\ref{noraml convolution}) becomes
\begin{equation} 
    y(l_0)=\sum^{N}_{i=1}\sum_{l_n \in R}w_i(l_n) \cdot x(l_0+l_n+d_i(l_0)) \label{one-d convolution}
\end{equation}

However, $d_i(l_0)$ usually contains a fraction part that can't be used to represent a pixel distance. Inspired by Linear Interpolation, arbitrary point can be represented by adjacent discrete points weighted by the distance.
\begin{equation}
x(l)=(l-\lfloor l \rfloor)\cdot x(\lceil l \rceil)+(\lceil l \rceil-l)\cdot x(\lfloor l \rfloor) \label{interpolation}
\end{equation}
where $l=l_0+l_n+d_i(l_0)$

Then Eq. (\ref{one-d convolution}) becomes
\begin{equation} 
    y(l_0)=\sum^{N}_{i=1}\sum_{l_n \in R}w_i(l_n) \cdot x(l) \label{advanced one-d convolution}
\end{equation}

As shown in Figure \ref{fig:dynamic}, the shape feature is obtained by Shape convolution on the input feature map with $N - 1$ output channels. While sampling, an OneDConv filter transforms its shape on each slide depending on the shape feature. For details of implementation, the pseudo-code of OneDConv network is shown in Algorithm \ref{alg:OneDConv Net}

\textbf{Complexity Analysis} OneDConv is expected to outperform normal 2D convolution with a small overhead over computation and parameters. The concept of Flops is proposed by NVIDIA \cite{molchanov2016pruning}, used to describe the quantity of floating-point operations. FLOPs for OneDConv is formulated as 
\begin{equation}
2HWC_{out}(C_{in}K^2+1)+2HW(K-1)(C_{in}K^2+1)
\end{equation}
where $H, W, C_{in}$ are height, width, and number of channels of the input feature map, $K$ is the kernel width(assumed to be symmetric), and $C_{out}$ is the number of output channels. Note that, the left part of the equation equals FLOPs of normal 2D convolution, while the rest of the equation is the overhead brought by Shape Convolution. Usually $C_{out}$ is greater than 16 , while K is set to 3 or 5 or 7. It's can be concluded that the additional overhead is negligible when compared to the origin FLOPs. Thus, OneDConv could be taken up without regard for its inefficiency in terms of time.

Furthermore, the number of extra parameters is also in a constant level. That's because one-dimension filters are derived from a 2D convolution filter, and Shape Convolution only requires a fixed number of filters.

\begin{algorithm}[ht]
\caption{One-dimension Convolution Networks}
\label{alg:OneDConv Net}
\begin{algorithmic}[1] %[1] enables line numbers
\STATE K $\leftarrow$ size of convolution filter
\STATE current epoch $\leftarrow$ 0, maximum epoch $\leftarrow$ 300
\STATE Initializing the parameters and network structure.
\WHILE{current epoch $\leq$ maximum epoch}
\STATE Inputting a mini-batch set of training images
\STATE In OneDConv operator, Producing shape feature map by forwarding shape convolution on the input feature map, using Eq. (\ref{shape convolution})
\STATE Executing K times One-dimension convolution and producing K output feature maps.
\STATE Integrating K output feature maps according to the shape feature.
\STATE Network forward convolution base on the input feature maps and transformed filters using Eq. (\ref{advanced one-d convolution}). And in the end connected by a fully connected layer.
\STATE Calculating the cross-entropy loss and then backward propagation.
\STATE Optimize the parameters
\STATE current epoch = current epoch + 1
\ENDWHILE
\end{algorithmic}
\end{algorithm}

\section{Experiments}

In this section, experiments are carried out on three popular benchmarks: MNIST \cite{lecun1998gradient}, CIFAR10 \cite{krizhevsky2009learning}, and ImageNette \cite{howard2020imagenette}. For each benchmark, following the common transform-invariant task settings, OneDConv is evaluated on origin and several types of distorted images, which are described in Sect. \ref{sec: experiment preparation}. In Sect. \ref{sec: mnist}, OneDConv is compared with several models on the MNIST using ResNet18 \cite{he2016deep}. In Sect. \ref{sec: cifar}, several models are evaluated on ResNet18 and PyramidNet110 \cite{DPRN} architectures. In Sect. \ref{sec: imagenette}, ImageNette, a subset of 10 classes from ImageNet with high resolution images, is used for experiments and visualization.

\subsection{Experiment Preparation}
\label{sec: experiment preparation}
\paragraph{Origin Dataset} The origin datasets are the raw images downloaded without any transformation. Many works only conduct experiments on distorted images while ignoring canonical images, which is incomplete because the real-world dataset contains many canonical images.

\paragraph{Rotated Dataset} The most common type of image transformation is rotation. In our experiment settings, the images are rotated with the angle ranging from $(-90^{\circ}, 90^{\circ})$, which is a common setting and is difficult enough to recognize. 

\paragraph{RTS Dataset} RTS is made up of the rotation, translation, and scale. In RTS distortion, the performance of the model is evaluated on the complicated composition of the transformation. In our experiment, the images are scaled by the random ratio ranging from $(0.7, 1)$. Then, the images are translated by the range of $(-5, 5)$ pixels in both horizontal and vertical directions, and rotated by the angles ranging from $(-45^{\circ}, 45^{\circ})$. The excess of the images is cropped during these transformations.

\paragraph{Experiment settings} To make a fair comparison, the same training settings are used for the hyperparameters. For the sake of training time balance, the models with the ResNet architecture are trained for $300$ epochs while the models with the PyramidNet architecture are trained for $200$ epochs in all the experiments. An SGD optimizer with a momentum of $0.9$ and weight decay of $5e-3$ is used to optimize all of the networks. Based on the number of parameters of models, the learning rate is tuned slightly to achieve the performance reported in each paper, which fluctuates from $0.05$ to $0.5$. To demonstrate transform-invariant ability while retaining comparative model performance, only simple data augmentation irrelevant to the distortion type is used, such as a random crop of images.

\subsection{Simple Digital Classification}
\label{sec: mnist}

\begin{table}
\centering
\setlength{\tabcolsep}{1.4mm}{
\begin{tabular}{llrrrr}
\toprule
\multirow{2}*{Method} & \multirow{2}*{Backbone} & \multirow{2}*{\#params(M)} & \multicolumn{3}{c}{\textbf{{Accuracy(\%)}}} \\
\cmidrule(lr){4-6}
~ & ~ & ~ & Origin & Rotated & RTS \\
\midrule
Vanilla  & \multirow{4}*{Resnet18}  & 11.17  & 98.99 & 96.29 & 98.17  \\
STN  &  ~ & 11.18 & 99.31 & 98.04 & 98.59  \\
GCN  & ~  & 15.46 & 99.36 & 97.87 & 98.50  \\
Ours & ~  & 11.24 & \bf99.41 & \bf98.17 & \bf98.87  \\
\bottomrule
\end{tabular}}
\caption{The experiments on MNIST}
\label{tab:MNIST}
\end{table}

The MNIST dataset consists of 60000 gray-scale images of $28\times28$ single digital numbers. All of the images are upsampled to $32 \times 32$ for the uniform network architecture. From the original MNIST and its variants MNIST-Rotated and MNIST-RTS, 50000 images are randomly selected for training and others for testing.

As comparison models to OneDConv, two typical models, Spatial Transformer Network and Gabor Convolution Network, were selected. The former contains a transform-invariant module the latter contains a transform-invariant filter. They are incorporated into the same architecture for a fair comparison, such as ResNet and PyramidNet shown in Table \ref{tab:network}, which reduces the influence of the number of parameters and network architectures.

From the experiment results in Table \ref{tab:MNIST}, it is found that our convolution outperforms the vanilla convolution and other models in both canonical and distorted images. Interestingly, models trained in MNIST appear to be more sensitive to rotation rather than RTS. For example, on the rotated dataset, performance on vanilla convolution is reduced by $2.7\%$, but only $0.82\%$ on the RTS dataset. It is possible that the original images in MNIST have already been scaled and rotated slightly, so the mild rotation and scaling in the RTS dataset have little effect on performance. Therefore, the models trained on MNIST are more robust to the RTS distortion. Compared to vanilla convolution, OneDConv improves $0.42\%$ in origin images, $1.88\%$ and $0.70\%$ in rotated and RTS images, respectively, outperforming other models, demonstrating a significant improvement in transform-invariant ability. 

\subsection{Nature Images Classification}
\label{sec: cifar}

\begin{table}
\centering
\setlength{\tabcolsep}{1mm}{
\begin{tabular}{llrrrr}
\toprule
\multirow{2}*{Method} & \multirow{2}*{Backbone} & \multirow{2}*{\#params(M)} & \multicolumn{3}{c}{\textbf{{Accuracy(\%)}}} \\
\cmidrule(lr){4-6}
~ & ~ & ~ & Origin & Rotated & RTS \\
\midrule
Vanilla  & \multirow{4}*{Resnet18}  & 11.17  & 91.96 & 79.88 & 79.01  \\
STN  &  ~ & 11.18 & 87.89 & 79.96 & 78.50  \\
GCN  & ~  & 15.46 & 88.26 & 79.32 & 79.26  \\
Ours & ~  & 11.24 & \bf92.31 & \bf81.33 & \bf80.05  \\
\midrule
Vanilla  & \multirow{4}*{PyramidNet}  & 28.49  & 96.01 & 85.50 & 85.55  \\
STN  &  ~ & 28.49 & 92.64 & 87.75 & 87.19 \\
GCN  & ~  & 43.24 & 92.66 & 87.77 & 86.43  \\
Ours & ~  & 29.30 & 95.88 & \bf88.71 & \bf88.94  \\
\bottomrule
\end{tabular}
}
\caption{The experiments on CIFAR10}
\label{tab:CIFAR}
\end{table}

\begin{figure}
    \centering
    \includegraphics[width=0.5\textwidth]{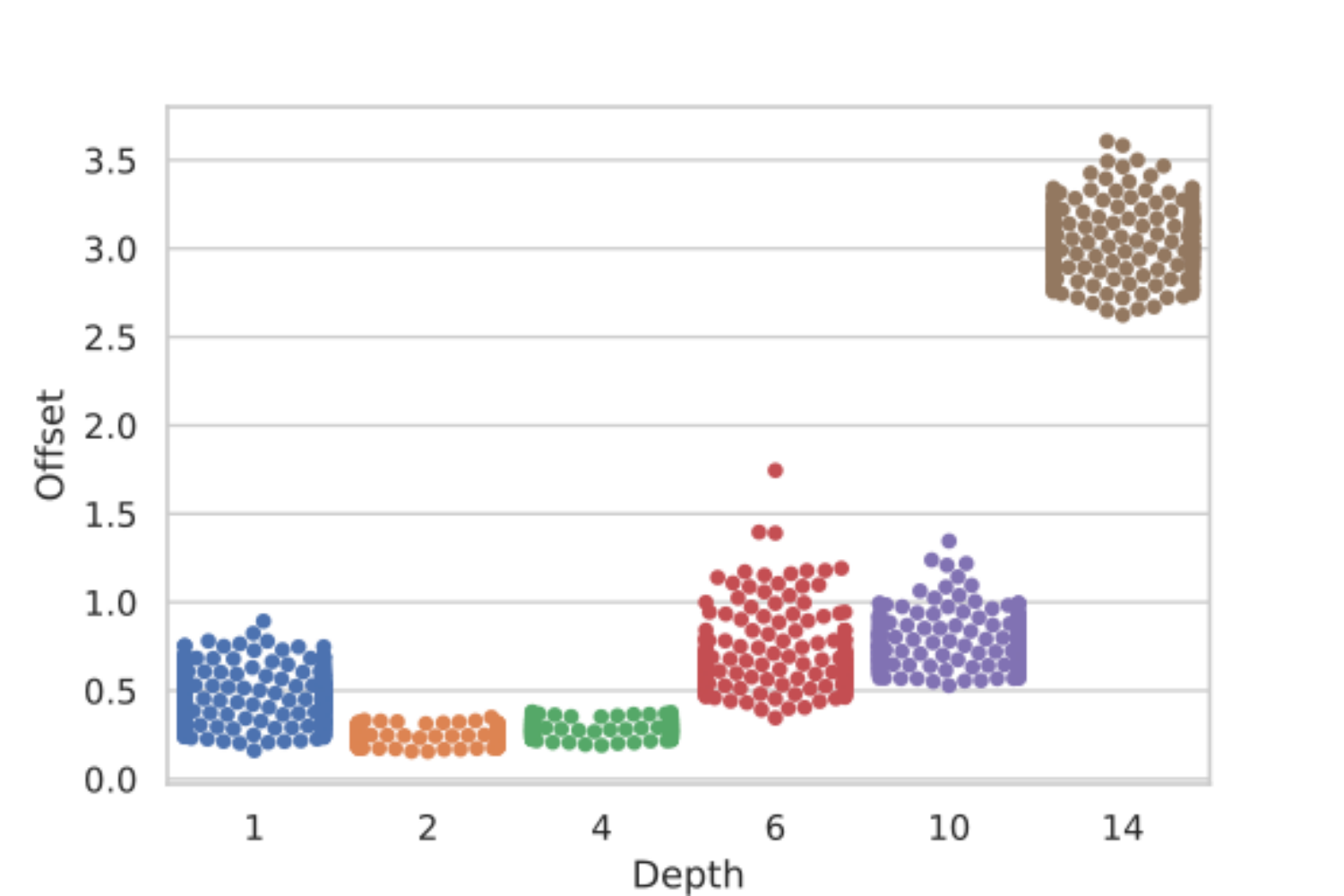}
    \caption{The maximum activation offsets are shown for the first convolution layer of each block in ResNet18. The offsets approach zero in the shallow layers, which allows the network to catch the low-level features similar to vanilla convolution. For deeper OneDConv layers, OneDConv tries to catch the semantic features following the distortion.}
    \label{fig:offset}
\end{figure}

Since MNIST is a small dataset with a relatively simple input structure, vanilla convolution has already been excellent in the MNIST dataset. The performance of OneDConv can not be demonstrated fully. 

In this section, experiments are based on a more complicated natural image dataset, CIFAR10. It is also a widely used visual classification benchmark, consisting of 60000 $32 \times 32$ real-world images divided into 10 classes. Similar to the previous dataset, the rotation and RTS transformation are applied to the original images, and the images are split into 50000 images for training and 10000 images for testing. 

In the experiments of CIFAR10 shown in Table \ref{tab:CIFAR}, two backbones are adopted, ResNet18 and PyramidNet110, to demonstrate the generation ability of the proposed convolution on different architectures. In PyramidNet, the widen factor $\alpha$ of Gabor convolution is set to $64$, while other models are set to $270$ with the consideration of parameter balance. Our model outperforms all the comparative models with both the ResNet18 and PyramidNet110 architectures. For origin images, the performance of all the comparative models drops around $3\%$ accuracy compared with vanilla convolution, indicating their weaker feature extraction ability on the nature images with complicated object structures. These models emphasize the invariance property while paying less attention to the feature extraction ability on the origin images. In contrast, our model does not sacrifice accuracy in the origin images when improving the performance of distorted images. Another thing to note is that in more powerful architectures like PyramidNet, OneDConv can outperform vanilla convolution by $3.21\%$ and $3.39\%$ on rotated and RTS images respectively, which far outweighs the improvement in ResNet18.

Furthermore, to show the insight of kernel distance in OneDConv, we visualize the offsets between vanilla convolution and OneDConv on the testset of CIFAR10-Rotated. The maximum offset activation for the first convolution layer of each block on ResNet18 is shown in Fig. \ref{fig:offset}. Interestingly, it is found that the shallower convolution layers try to catch the low-level feature with small transformations just like vanilla convolution, while deeper layers are more likely to transform the kernel with higher offsets.

\subsection{Feature Extraction on High Resolution Images}
\label{sec: imagenette}

\begin{figure*}[htbp]  
\centering  
\includegraphics[width=0.8\textwidth]{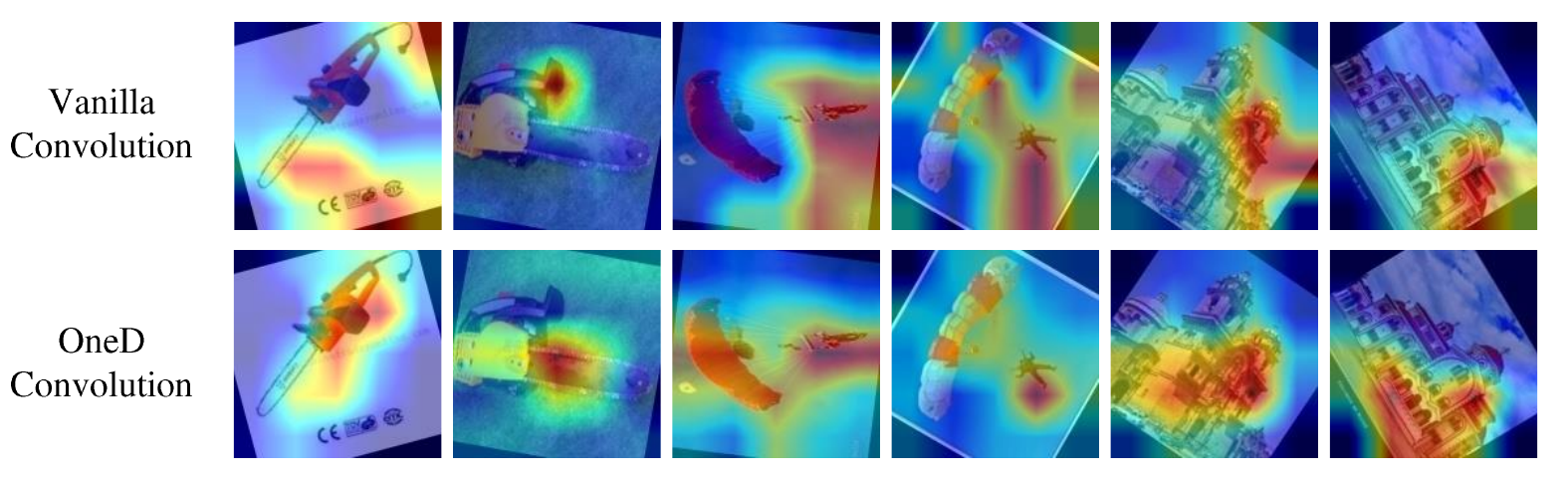}  
\caption{The heat map produced by Grad-CAM shows the important regions in the images that are marked in red. It is illustrated that when the images are distorted, the OneDConv network extracts the feature effectively, especially for entities that require a larger receptive field, whereas vanilla convolution fails.}
\label{fig:cam}
\end{figure*}

\begin{table}
\centering
\setlength{\tabcolsep}{1.4mm}{
\begin{tabular}{llrrrr}
\toprule
\multirow{2}*{Method} & \multirow{2}*{Backbone} & \multirow{2}*{\#params(M)} & \multicolumn{3}{c}{\textbf{{Accuracy(\%)}}} \\
\cmidrule(lr){4-6}
~ & ~ & ~ & Origin & Rotated & RTS \\
\midrule
Vanilla  & \multirow{3}*{Resnet50} & 23.52 & 88.71 & 76.51 & 77.41  \\
GCN  &  ~ & 27.93 & 87.75 & 78.10 & 77.91   \\
Ours & ~  & 23.59 & \bf89.96 & \bf78.52 & \bf79.46  \\
\bottomrule
\end{tabular}}
\caption{The experiments on ImageNette}
\label{tab:ImageNette}
\end{table}

In the previous sections, we discussed the performance of various models and architectures in the context of small images, which is not enough to prove OneDConv can solve practical. In this section, the ability of feature extraction to work on higher resolution images will be analyzed. For this purpose, ImageNette is selected as our third benchmark, which is a subset of 10 classified classes from ImageNet with a resolution of $160 \times 160$. 

The experimental result is shown in Table \ref{tab:ImageNette}. Since the localisation module of STN is difficult to adapt to the high resolution images, its performance is poor on this benchmark. For this reason, its result is ignored in our experiment table. For canonical images, OneDConv outperforms the vanilla convolution by $1.25\%$ whereas the Gabor convolution fails. The performance on rotated images is similar between OneDConv and Gabor convolution. But in RTS images, the accuracy of OneDConv is ahead of vanilla convolution by $2.05\%$ and Gabor convolution by $1.55\%$.

In Fig. \ref{fig:cam}, it's the heat map of classification results for vanilla convolution and  OneDConv, using the visualization method Gradient-weighted Class Activation Mapping (Grad-CAM) \cite{selvaraju2017grad}. It is observed that vanilla convolution is more difficult to extract the features due to the need for a larger receptive field on distorted images, while OneDConv can recognize the object correctly due to the flexible spatial filters.

\section{Conclusion}
In this paper, we propose a novel one-dimensional convolution operator, {\it OneDConv}, which can replace the vanilla convolution to allow CNN to extract transform-invariant visual representations. In both origin and distorted images, our model outperforms the vanilla convolution and comparative models. With OneDConv, the heavy reliance on data quality and data augmentation can be greatly reduced, allowing CNN to be applied to more complex real-world scenarios. In future work, more architectures using OneDConv can be evaluated in other vision tasks, such as object detection and semantic segmentation, with more robust internal transform-invariant mechanisms supported.

% \section*{Acknowledgements}
% We sincerely thank the reviewers who participated in the peer review process for this paper. Also thanks to everyone for the discussion and helpful advice about OneDConv.

\bibliographystyle{named}
\bibliography{ijcai22}
\end{document}